\documentclass[11pt]{article}

\usepackage[utf8]{inputenc}
\usepackage[T1]{fontenc}
\usepackage{lmodern}
\usepackage{graphicx}
\usepackage{amsmath}
\usepackage{booktabs}
\usepackage{enumitem}
\usepackage[margin=1in]{geometry}
\usepackage{hyperref}

\usepackage{longtable}
\usepackage{graphicx} 
\usepackage{tcolorbox}
\usepackage{xcolor}

\definecolor{maincolor}{RGB}{0,102,204} 

\begin{document}
\title{IMPACTeen: Intentions, Manipulation, Persuasion, Annotations, and Consequences in Teen Communication Dataset}

\author{
Aleksander Szczęsny\footnote{Corresponding author: \href{mailto:aleksander.szczesny@pwr.edu.pl}{aleksander.szczesny@pwr.edu.pl}},
Wiktoria Mieleszczenko-Kowszewicz,
Maciej Markiewicz,\\
Beata Bajcar,
Tomasz Adamczyk,
Jolanta Babiak,
Grzegorz Chodak,\\
Przemysław Kazienko\\[1ex]
Wrocław University of Science and Technology, Wrocław, Poland
}

\date{}

\maketitle

\begin{abstract}
IMPACTeen is a dataset of textual social influence scenarios spanning interpersonal, media-based, and digital settings in an adolescent context. It contains 1,021 texts, 5,100 individual annotation records, and gold labels for social influence techniques, with each text annotated from five distinct perspectives: teenagers, parents, psychologists, communication experts, and teachers. The resource was constructed through constrained LLM generation, followed by a two-step human editing and validation phase aimed at ensuring youth-context realism. A multi-dimensional annotation covered influence presence, techniques, intentions, consequences, resistance, reactions, and annotation confidence. The dataset supports research on social influence detection, annotator disagreement, cross-lingual modeling, and the training and evaluation of language models. The dataset was created in Polish and is accompanied by a corresponding English version.
\end{abstract}

\section{Background \& Summary}
\textit{IMPACTeen} (Intentions, Manipulation, Persuasion, Annotations, and Consequences in Teen communication) is a dataset of texts, including dialogues, depicting social influence among adolescents. Each text represents a situation in which one party attempts to influence the attitudes, beliefs, or behaviors of a young individual or a group. The thematic and situational scope is wide, spanning interpersonal, media-based, and digital communication settings, as well as a variety of contexts and domains. The texts were artificially generated under controlled conditions and underwent a two-stage process of human editing and expert validation to ensure youth-context realism. They were manually annotated in multiple dimensions, including the presence, form, intentions, reactions, and consequences of social influence, described in detail in the following sections.

Resources of this kind are scarce. There are few datasets that describe social influence in interpersonal communication \cite{wang2024mentalmanip, wang2019pfg, zhang2026persuasiondoubleblindmultidomaindialogue, young2011, tan2016, mieleszczenko2025sitt}, but none that specifically target adolescents and young people, a group particularly exposed to such influence. Furthermore, existing resources rarely record judgments from several distinct annotator perspectives. For Polish in particular, no comparable resource exists. \textit{IMPACTeen} addresses these gaps by providing parallel Polish and English versions of the data, accompanied by annotations from five distinct annotator groups: parents, teenagers, psychologists, communication experts, and teachers. A subset of texts was annotated twice---at the beginning and at the end of the annotation process---to enable the assessment of annotation stability and annotator competence development over time in a study conducted in parallel \cite{markiewicz2026annotation}.

The main value of the dataset lies in the combination of (i) controlled, context-driven generation of texts, (ii) manual correction of the generated material to improve youth-context realism, (iii) multi-dimensional annotation of social influence, and (iv) the involvement of five distinct annotator groups. The dataset can support future interdisciplinary research, such as the analysis and detection of social influence, the study of social influence subjectivity among annotator groups, the training and evaluation of Large Language Models, and the assessment of dataset diversity.

\begin{figure}[htbp]
  \centering
  \includegraphics[width=\textwidth, trim={0 39cm 0 0},clip]{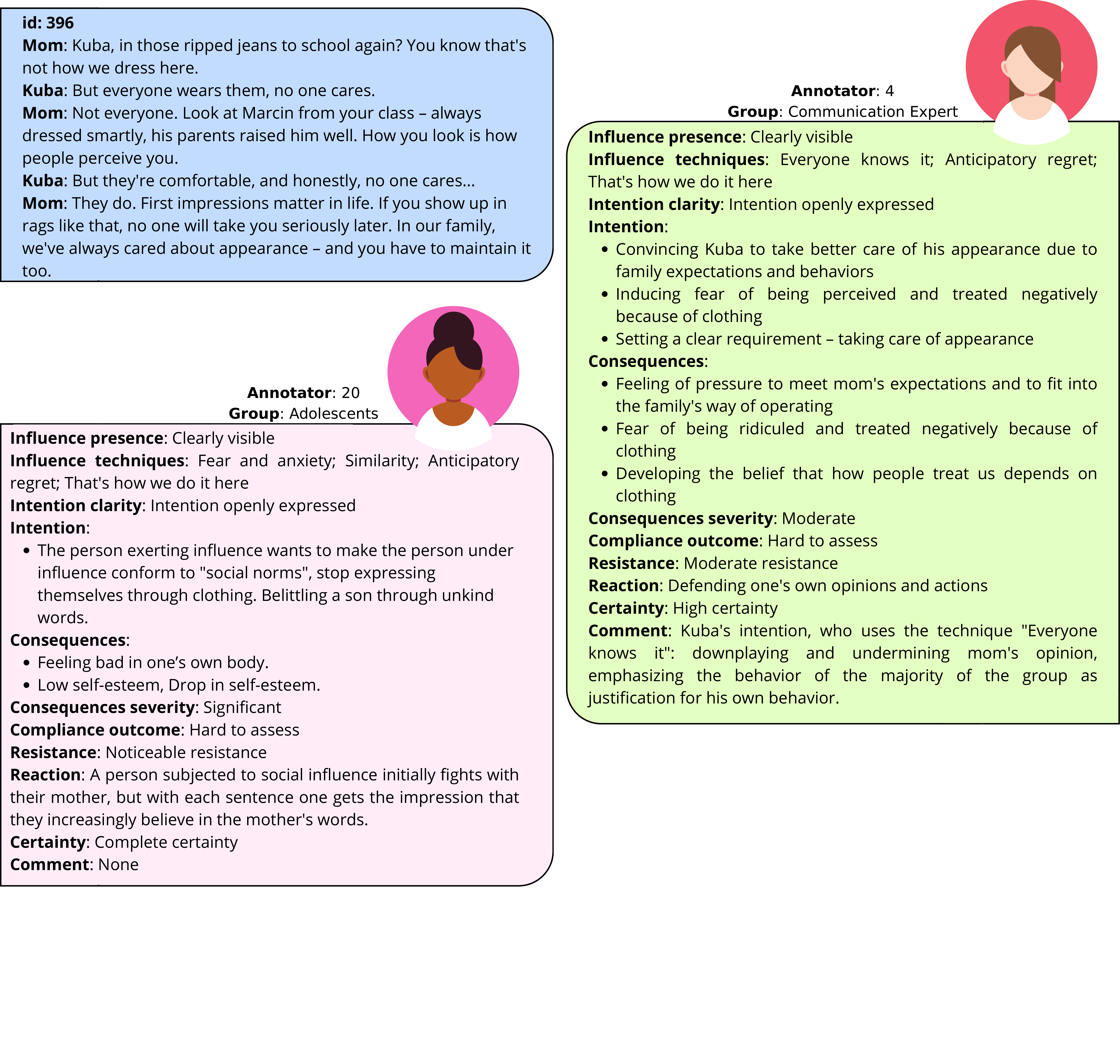}
  \caption{A sample text from the dataset with two (out of five) individual annotations.}
  \label{fig:annotation_example}
\end{figure}

An example text can be seen in Figure~\ref{fig:annotation_example}.

\section{Methods}

\subsection{Dataset construction}
\begin{figure}[htbp]
  \centering
  \includegraphics[width=\textwidth, trim={0 1.3cm 0 0},clip]{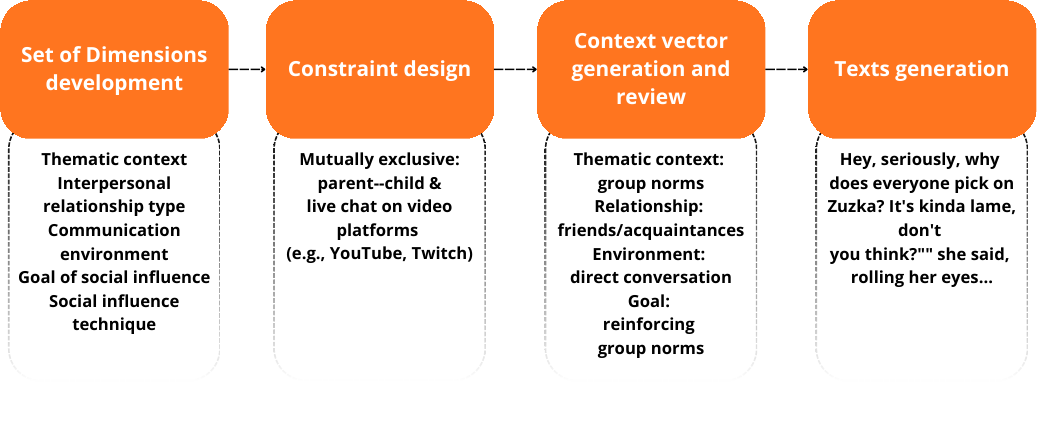}
  \caption{Texts generation pipeline.}
  \label{fig:process}
\end{figure}

The dataset was developed through a multi-stage pipeline, Figure~\ref{fig:process}, consisting of the following stages:
\begin{itemize}
    \item literature-based selection of techniques of social influence that should be the most prominent in adolescent communication from the SITT taxonomy \cite{mieleszczenko2025sitt}, serving as a theoretical basis,
    \item development of a set of dimensions representing social influence in the interactions of teenagers, including: thematic context, type of interpersonal relationship, setting and platform, goal of social influence, discretion of the applied technique, suggested length, recipient's resistance to social influence, and the social influence technique,
    \item identification of co-occurring and mutually exclusive dimensions (for example, parent--child influence in live chat on video platforms),
    \item manual review and correction of \textit{context vectors},
    \item generation of texts based on the validated \textit{context vectors}.
\end{itemize}

\subsubsection{Social influence techniques selection}

We used the SITT taxonomy \cite{mieleszczenko2025sitt} as a base of social influence techniques. The set of 20 social influence techniques was selected because these techniques are plausible and developmentally relevant in adolescent communication. Adolescents are especially exposed to influence based on peer norms, belonging, social comparison, approval, fear of exclusion, authority, emotional pressure, and identity-related appeals. All selected techniques are listed in Table~\ref{tab:taxonomy} under the \textit{Social influence technique} dimension. These techniques allow the dataset to represent both explicit and subtle forms of social influence that may appear in everyday adolescent interactions.

\subsubsection{A set of dimensions for context vectors}

To ensure the systematic and semantic diversity of the synthetic dialogues, we defined a set of dimensions intended to differentiate the generated conversations along several dimensions, which are presented in Table~\ref{tab:taxonomy}. It was designed to capture both the situational context of influence and the communicative forms through which influence may be expressed in interactions involving  adolescents and young people. This structure made it possible to generate texts that were diverse while remaining comparable across cases. The inclusion of the type of interpersonal relationship and communication environment was particularly important because adolescent social influence often occurs in different relational settings, including parent–child, teacher–student, peer, romantic and incidental interactions, as well as in face-to-face, digital, media-based and written communication contexts.

A \textbf{context vector} is a single combination of the set of dimensions. Each example was generated using a constrained generation approach, in which the model is prompted with predefined attribute values and compatibility constraints \cite{evuru2024coda, fu2022effective}. The elements that varied across generated examples were the values assigned to the dimensions, and a single set of values is referred to as the \textit{context vector}. For some dimensions, such as \textit{Communication environment}, the possible values are additionally grouped into broader categories. A full description of all dimensions and their possible values is provided in Table~\ref{tab:taxonomy}.

Due to the nature of the task, certain values of specific dimensions impose dependencies on values in other dimensions, whereas some value combinations were mutually exclusive. These relationships were determined based on the authors' expert knowledge and commonsense judgment, formalized as constraint rules, and are presented in Table~\ref{tab:excluded_feature_combinations} and Table~\ref{tab:required_feature_combinations}.

Using this framework, 330 \textit{context vectors} were generated, which were subsequently manually reviewed and validated by a social influence expert. Each validated \textit{context vector} was then used to generate text examples. 

\begin{longtable}{p{0.23\textwidth} p{0.72\textwidth}}
\caption{Set of dimensions used to guide generation and annotation of social influence scenarios.}
\label{tab:taxonomy} \\

\hline
\textbf{Dimension} & \textbf{Values} \\
\hline
\endfirsthead

\hline
\textbf{Dimension} & \textbf{Possible values} \\
\hline
\endhead

\hline
\endfoot

Thematic context &
Distribution of commercial/consumer goods; gadget-oriented consumption; music creation or interest in music (including music bands); films; advertisements; sports; school; cosmetics and brand prestige; image-related pressure; group norms and group belonging. \\

Interpersonal relationship type &
\textbf{Hierarchical relationships:} teacher--student; parent--child; relationship with an authority figure. \newline
\textbf{Peer and romantic relationships:} romantic relationships; friends/acquaintances. \newline
\textbf{Casual relationships:} incidental or one-off interactions. \\

Communication environment &
\textbf{Face-to-face interactions:} dialogue and conversation transcripts; letters and e-mails; SMS and mobile communication. \newline
\textbf{Print media:} press articles (including interviews, news, reviews, columns, and opinion pieces); advertisements and sponsored materials; feature reports and investigative journalism. \newline
\textbf{Radio and television:} interviews; news; sponsored materials and television advertisements; popular science programs; entertainment programs; films; feature reports and investigative journalism. \newline
\textbf{Digital environments:} original social media posts (e.g., Facebook, Instagram, TikTok); single social media comments; social media discussions, including internet forums, Reddit, and interest groups; messaging platforms (e.g., WhatsApp, Telegram); live chat on video platforms (e.g., YouTube, Twitch); communication between players in online games. \\

Goal of social influence &
\textbf{Changing attitudes and beliefs:} worldview conversion; opinion shaping. \newline
\textbf{Behavioral control:} encouraging specific actions; discouraging action. \newline
\textbf{Building group identity:} constructing an ``us vs.\ them'' divide; reinforcing group norms. \newline
\textbf{Gaining material and non-material benefits:} increasing profits; acquiring non-material resources. \newline
\textbf{Preserving the status quo:} maintaining an existing position. \newline
\textbf{Image-building:} building the image of the author, another person, or a product/brand. \\

Social influence technique &
Anticipatory regret; Humor; Fear and anxiety; Guilt; The power of word “love”; Emotional see-saw; Show disappointment; Liking; Similarity; Flattery; Authority of person or science; Everyone knows it; The “We” rule; We are exceptional; That's how we do it here; We are looking for people like you; Gratitude; Give to take; Door-in-the-face; Labeling. \\

Technique visibility &
Subtle and discreet; obvious and explicit. \\

Text length &
A single sentence, a single utterance, or a short non-dialogic text; a short dialogue (2--4 turns); a medium-length dialogue (4+ turns). \\

Target resistance outcome &
The target complies easily; the target resists but ultimately yields; the target does not yield; it is unclear whether the target yields. \\

\end{longtable}

{
\begin{longtable}{p{0.30\textwidth} p{0.65\textwidth}}
\caption{Feature combinations that should not be used together.}
\label{tab:excluded_feature_combinations} \\
\hline
\textbf{Feature} & \textbf{Do not combine with} \\
\hline
\endfirsthead
\hline
\textbf{Feature} & \textbf{Do not combine with} \\
\hline
\endhead
\hline
\endfoot
parent--child &
live chat on video platforms (e.g., YouTube, Twitch); communication between players in online games \\
teacher--student &
live chat on video platforms (e.g., YouTube, Twitch); communication between players in online games; increasing profits; building the image of a product/brand \\
romantic relationships &
live chat on video platforms (e.g., YouTube, Twitch) \\
friends/acquaintances &
live chat on video platforms (e.g., YouTube, Twitch) \\
interviews &
a single sentence, a single utterance, or a short non-dialogic text \\
dialogue and conversation transcripts &
a single sentence, a single utterance, or a short non-dialogic text \\
social media discussions, including internet forums, Reddit, and interest groups &
a single sentence, a single utterance, or a short non-dialogic text \\
communication between players in online games &
a single sentence, a single utterance, or a short non-dialogic text \\
cosmetics and brand prestige &
teacher--student \\
\end{longtable}
}

{
\begin{longtable}{p{0.30\textwidth} p{0.65\textwidth}}
\caption{Feature combinations that should be connected.}
\label{tab:required_feature_combinations} \\
\hline
\textbf{Feature} & \textbf{Should be combined with} \\
\hline
\endfirsthead
\hline
\textbf{Feature} & \textbf{Should be combined with} \\
\hline
\endhead
\hline
\endfoot
original social media posts (e.g., Facebook, Instagram, TikTok) &
a single sentence, a single utterance, or a short non-dialogic text \\
news &
a single sentence, a single utterance, or a short non-dialogic text \\
live chat on video platforms (e.g., YouTube, Twitch) &
relationship with an authority figure; incidental or one-off interactions \\
single social media comments &
a single sentence, a single utterance, or a short non-dialogic text \\
a single sentence, a single utterance, or a short non-dialogic text &
it is unclear whether the target yields \\
\end{longtable}
}

\subsubsection{Texts generation}

For each validated \textit{context vector}, three distinct texts were generated using the prompt presented in Section~\ref{prompt:generate_text_pl}. Using the validated set of 330 context vectors, we obtained 990 initial examples. In addition, 100 texts were further adapted into narrative-style scenarios in order to enrich contextual detail and increase variation in discourse form. Furthermore, based on another 40 randomly selected texts, one additional example without social influence was generated for each selected text, thereby introducing negative instances into the dataset. After quality control, nine texts were removed, resulting in a final dataset of 1,021 texts. All texts were generated with the \textit{DeepSeek-V3} model with temperature set at 1.0.

During dataset construction, we aimed to maintain a reasonably balanced distribution of social influence techniques across the generated texts. In practice, some techniques proved easier for the model to generate than others during the preliminary inspection. As a result, the distribution of techniques across the final set of context vectors was not uniform.

Another important factor was that the model was not constrained to express only a single social influence technique. Although each text was generated to reflect one target technique, additional techniques could also appear in the output if the model introduced them. This possibility was also explicitly allowed in the generation prompt (see Section~\ref{prompt:generate_text_pl}).

\subsection{Annotator characteristics}
The annotation process involved 25 participants, recruited to represent five distinct stakeholder groups relevant to the study, with five annotators per group: (1) teenagers, (2) parents of at least two teenagers, (3) experts from the communication industry, (4) psychologists, and (5) teachers. This balanced design was intended to capture a diverse range of perspectives on the annotated content, reflecting both the experiential viewpoint of adolescents and their parents, as well as the professional expertise of communication specialists, psychologists, and educators. The annotator group comprised 11 men and 14 women, with a mean age of 34.2 years (SD = 11.36).

\subsection{Annotation procedure}
The annotation process was divided into two phases: (1) text validation and manual adjustment to improve youth-context realism, and (2) multi-dimensional annotation of social influence.

\textbf{Phase 1: text validation and manual adjustment.} 
In the first phase, annotators assessed the ecological validity of the text. Each annotator was presented with a text involving young people, either among themselves or with people from their immediate environment (parents, school staff, peers, or other adults). The annotators' task was to read the text and evaluate whether it realistically reflected typical communication patterns of adolescents.
If a text was judged to be representative of authentic youth communication, the annotator accepted it without modification and proceeded to the next item. However, if the text was perceived as unnatural, the annotator was instructed to edit the text to make it more realistic and natural for a youth context. Importantly, annotators were explicitly instructed to avoid removing larger portions of the text in order to preserve the social influence techniques embedded in the text. During this process, 563 texts were adjusted.

Because the texts could be modified during this phase, and each text could be independently revised by two annotators, experts manually evaluated the resulting versions. For each case, one of three versions was selected: the original text, the version corrected by the first annotator, or the version corrected by the second annotator. The selection criteria included linguistic correctness, grammatical accuracy, stylistic quality, and preservation of the original social influence in the text. If none of the three versions were considered satisfactory, the text was revised again. Nine texts were finally removed from the final dataset, as they did not meet the required quality standard even after additional revision.

\textbf{Phase 2: Multi-level annotation.} In the second phase, the validated texts were annotated along ten dimensions capturing the character and consequences of social influence:
\begin{enumerate}
\item \textit{Influence presence} -- rated on 5-point Likert-scale (1 -- No influence to 5 -- Unambiguous), with an additional \textit{hard to assess} option.
\item \textit{ Identification of social influence techniques} -- Identification of social influence techniques – annotators identified the social influence techniques present in the text based on a predefined taxonomy of 20 techniques. A given dialogue could contain one, several, or none of the listed techniques. 
\item  \textit{Assessment of intention clarity}. -- assessed on a 5-point Likert scale (1 -- No visible intention to 5 -- Intention openly expressed) with an additional \textit{hard to assess} option, indicating how explicit the influencer’s intention was.
\item \textit{Recognition of the influencer's intention} -- open-ended textual description of what might be the intention of the influencer.
\item \textit{Potential consequences of being influenced}  -- open-ended textual description of the possible consequences of being influenced. That covers both immediate effects and longer-term repercussions concerning emotions, decisions, relationships, and the target's life or social situation.
\item \textit{Severity of consequences}. -- rated on 5-point Likert-scale (1 -- Minimal/none to 5 -- Serious) with an additional \textit{hard to assess} option.
\item \textit{Compliance outcome.} -- rated using two substantive responses (\textit{yielded} vs. \textit{did not yield}) with an additional \textit{hard to assess} option. Annotators assessed whether the target person yielded to the influence attempt.
\item \textit{Degree of resistance to influence} -- assessed using a 5-point Likert-scale (1 -- No resistance to 5 -- Very strong resistance) with an additional \textit{hard to assess} option.
\item \textit{Reaction to influence} -- free-text description of the target person’s reaction to the influence attempt.
\item \textit{Annotation confidence}  -- annotators rated on a 5-point Likert scale (1 -- No certainty at all to 5 -- Complete certainty) their self-confidence in evaluating the scenarios in all previous questions. This rating was also required in cases where the social influence technique was not identified in the text.
\end{enumerate}

Annotators were instructed not to complete the annotation besides \textit{Annotation confidence} if they assessed that the text did not contain any form of social influence. Despite this instruction, some annotations contain missing values, including cases in which the text was annotated as containing social influence. Details are presented in Table~\ref{tab:missing_hard_to_assess}.

\subsubsection{Repeated-annotation design.}
To assess annotation stability over time in phase 2, each annotator first completed annotations for an initial set of 30 texts. This was followed by the main annotation batch of 174 or 175 additional texts. After completion of the main batch, access to the earlier annotations was removed, and the same initial set of 30 texts was presented again for repeated annotation. This design made it possible to compare initial and repeated judgments for the same annotator and the same texts \cite{markiewicz2026annotation}.

At the dataset level, each text was intended to receive one annotation from each annotator group. However, within each group, text assignments were randomized independently for individual annotators rather than matched across annotators. As a result, there was no fixed overlap pattern whereby the texts annotated by one annotator were necessarily annotated by the same annotator in another group.

\begin{longtable}{p{0.78\textwidth} p{0.16\textwidth}}
\caption{Annotation questions and their short labels.}
\label{tab:annotation_questions} \\
\hline
\textbf{Question} & \textbf{Short label} \\
\hline
\endfirsthead
\hline
\textbf{Question} & \textbf{Short label} \\
\hline
\endhead
\hline
\endfoot
To what extent is social influence present? &
\rule{0pt}{5ex}\shortstack{\textit{Influence} \\ \textit{presence}} \\
Which social influence techniques are present in the text? &
\rule{0pt}{5ex}\shortstack{\textit{Influence} \\ \textit{techniques}} \\
How explicit is the intention of the person exerting influence? &
\rule{0pt}{5ex}\shortstack{\textit{Intention} \\ \textit{clarity}} \\
Indicate what intentions the person exerting influence may have had, as well as their goals or expectations toward the recipient. If you identify several intentions, enter them on separate lines. Use sentence fragments where possible. If you do not identify any intention, leave this field blank. &
\textit{Intention} \\
What may be the consequences of yielding to social influence? Include both direct effects and more distant consequences related to emotions, decisions, relationships, life situation, or social situation. If you identify several consequences, enter them on separate lines. Use sentence fragments where possible. If you do not identify any consequences, leave this field blank. &
\textit{Consequences} \\
How significant might the effects or consequences of yielding to the influence be for the person yielding to it? &
\rule{0pt}{1ex}\shortstack{\textit{Consequences} \\ \textit{severity}} \\
Did the person subjected to social influence yield to it, or did they remain independent? &
\rule{0pt}{1ex}\shortstack{\textit{Compliance} \\ \textit{outcome}} \\
How strongly did the person subjected to social influence resist it? &
\textit{Resistance} \\
What was the reaction of the person subjected to social influence? If you identify several reactions, enter them on separate lines. Use sentence fragments where possible. If you do not identify any reactions, leave this field blank. &
\textit{Reaction} \\
Please assess the degree of certainty of your annotation regarding the above answers; also in the case where no influence technique was selected. & 
\textit{Annotation confidence} \\
\end{longtable}

\subsection{Inter-annotator agreement}

Inter-annotator agreement was assessed at text level using metrics matched to the measurement level of each annotation variable. For each text, all pairs of available annotations were compared. Responses marked as \textit{hard to assess} were treated as unavailable for the corresponding variable and excluded from agreement computations.

For ordinal variables (\textit{influence presence}, \textit{intention clarity}, \textit{consequences severity}, and \textit{resistance}), agreement was summarized using pooled pairwise quadratic weighted Cohen's kappa, which is appropriate for ordered categories because it penalizes larger disagreements more strongly than smaller ones \cite{cohen1960agreement, mcgrath2010kappa, li2023kappa}.  All ordered annotation dimensions were mapped to integer values that preserved the rank order of the original response categories, starting from 0 and increasing by 1. The option \textit{hard to assess} was not coded as an ordinal value in agreement computation. For the nominal variable \textit{compliance outcome}, agreement was summarized using pooled pairwise Cohen's kappa \cite{cohen1960agreement}. For the multi-label variable \textit{techniques}, agreement was measured using Jaccard similarity \cite{jaccard1912flora}. For the free-text variables (\textit{intention}, \textit{consequences}, and \textit{reaction}), agreement was approximated using cosine similarity between precomputed text embeddings. Similarity was computed only for texts with at least two non-empty responses for the corresponding variable. For Polish data, the \textit{sdadas/mmlw-roberta-large} model \cite{dadas2024pirb} was used to generate embeddings. Similarly, for English data, the \textit{microsoft/deberta-v3-base} model \cite{he2021deberta} was used.

\subsection{Gold labels}

Due to the nature of the annotation scheme, the gold labels were assigned only to the variable of the \textit{social influence technique}. The initial gold labels were derived by majority vote, with a label assigned to a text when it was selected by at least three out of five annotators. In cases where no technique met this threshold, the final label was determined through expert adjudication by a psychologist. A total of 84 texts were processed in this way.

\subsection{Final dataset preparation}

To prepare the released dataset and analysis files, we aligned the generation contexts, the final text versions used in annotation, and the annotation records using the shared \texttt{id} field. The English versions were machine-translated using the prompt presented in Section~\ref{prompt:translate_pl}. The same \textit{id} value was preserved across language versions. The social influence techniques were modeled as a multi-label categorical variable, because each dialogue could contain multiple techniques simultaneously or none at all. In the released files, the identified techniques were stored in textual form as labels separated by newline characters. Textual annotations were not reduced to numeric scores. Instead, free-text fields were preserved as original text, allowing them to be analyzed separately as qualitative or linguistic data. The option \textit{hard to assess} was preserved.

\subsection{Ethics statement}

Ethical approval was obtained for the participation of adult annotators in the study. The study did not involve research with child participants and did not include the collection of natural or private conversations with minors. All annotations in the \textit{teenager} group were provided by participants aged 18 to 19 years. The dataset consists of generated and manually edited texts representing social influence situations involving adolescent characters rather than real adolescents' interactions. No private real-world conversations were recorded, collected, or reproduced. All participants provided their informed consent prior to participation.

\section{Data Records}

The released resources consist of parallel Polish and English files describing the same set of generated scenarios, together with annotations, repeated-annotation comparisons, and gold labels. An overview of the released files is provided in Table~\ref{tab:released_files_overview}.

\begin{longtable}{p{0.28\textwidth} p{0.10\textwidth} p{0.12\textwidth} p{0.42\textwidth}}
\caption{Released dataset files.}
\label{tab:released_files_overview} \\

\hline
\textbf{File name} & \textbf{Format} & \textbf{Records} & \textbf{Description} \\
\hline
\endfirsthead

\hline
\textbf{File name} & \textbf{Format} & \textbf{Records} & \textbf{Description} \\
\hline
\endhead

\hline
\endfoot

all\_annotations\_pl.csv & CSV & 5100 & Full set of annotation records in Polish, including the repeated annotation of 30 texts. \\
all\_annotations\_en.csv & CSV & 5100 & Full set of annotation records in English, including the repeated annotation of 30 texts. \\
\shortstack[l]{repeated\_annotation\_\\comparison\_pl.csv} & CSV & 749 & Polish file containing side-by-side initial and repeated annotations for the re-annotated texts. \\
\shortstack[l]{repeated\_annotation\_\\comparison\_en.csv} & CSV & 749 & English file containing side-by-side initial and repeated annotations for the re-annotated texts. \\
gold\_labels\_pl.csv & CSV & 1021 & Final gold social influence technique labels in Polish, including 10-fold assignments and expert annotation status. \\
gold\_labels\_en.csv & CSV & 1021 & Final gold social influence technique labels in English, including 10-fold assignments and expert annotation status. \\
generation\_contexts\_pl.csv & CSV & 1030 & Polish generation contexts and generated texts, including records later removed during quality control. \\
generation\_contexts\_en.csv & CSV & 1030 & English generation contexts and generated texts, including records later removed during quality control. \\
\end{longtable}

The Polish and English versions of the released files are aligned with \textit{the id} value; the same \texttt{id} value refers to the same scenario across all the provided files, while the language-specific columns store the corresponding Polish or English textual content and labels.

\subsection{Annotation files}

The main dataset is the \textit{all\_annotations\_[pl|en].csv} file, which contains the complete set of annotation records for all texts. The annotation files preserve the original multi-perspective structure of the dataset. Each record corresponds to one annotation instance for one text, provided by one annotator, and includes the full annotation together with the text content and annotator-group information. The schema of the annotation files is described in Table~\ref{tab:all_annotations_columns}. 

\begin{longtable}{p{0.34\textwidth} p{0.14\textwidth} p{0.44\textwidth}}
\caption{Columns in the all-annotations files.}
\label{tab:all_annotations_columns} \\

\hline
\textbf{Column name} & \textbf{Type} & \textbf{Description} \\
\hline
\endfirsthead

\hline
\textbf{Column name} & \textbf{Type} & \textbf{Description} \\
\hline
\endhead

\hline
\endfoot

id & identifier & Unique text identifier. \\
annotator & identifier & Annotator identifier. \\
group & categorical & Annotation batch or group identifier. \\
text\_en / text & text & Annotated text instance. \\
question\_1\_influence\_presence & ordinal & Influence presence rating; see Table~\ref{tab:annotation_questions}. \\
question\_2\_techniques & multilabel & Social influence techniques identified in the text; see Table~\ref{tab:annotation_questions}. \\
question\_3\_intention\_clarity & ordinal & Clarity of the influencer's intention; see Table~\ref{tab:annotation_questions}. \\
question\_4\_intention & text & Free-text description of the influencer's intention; see Table~\ref{tab:annotation_questions}. \\
\shortstack[l]{question\_5\_\\consequences} & text & Free-text description of possible consequences; see Table~\ref{tab:annotation_questions}. \\
\shortstack[l]{question\_6\_\\consequences\_severity} & ordinal & Severity of consequences; see Table~\ref{tab:annotation_questions}. \\
question\_7\_submission & ordinal & Compliance outcome; see Table~\ref{tab:annotation_questions}. \\
question\_8\_resistance & ordinal & Degree of resistance to influence; see Table~\ref{tab:annotation_questions}. \\
question\_9\_reaction & text & Free-text description of the target's reaction; see Table~\ref{tab:annotation_questions}. \\
question\_10\_certainty & ordinal & Annotation confidence; see Table~\ref{tab:annotation_questions}. \\
comment & text & Optional annotator comment. \\
\end{longtable}

\subsection{Repeated annotation comparison files}

The \textit{repeated\_annotation\_comparison\_[pl|en].csv} files are designed to support research on differences in annotation skills over time. They contain paired initial and repeated annotations for the subset of texts that were re-annotated at the beginning and end of the annotation process. The schema of these files is described in Table~\ref{tab:repeated_annotations_columns}. 

The files contain 749 paired records rather than 750 because one repeated-annotation pair was incomplete and was excluded from the released comparison files.

\begin{longtable}{p{0.30\textwidth} p{0.14\textwidth} p{0.48\textwidth}}
\caption{Schema of the repeated-annotation comparison files. The Polish and English versions share the same schema. Metadata columns occur once, whereas each annotation field is present twice: once with the prefix \texttt{initial\_annotation\_} and once with the prefix \texttt{repeated\_annotation\_}.}
\label{tab:repeated_annotations_columns} \\

\hline
\textbf{Base column name} & \textbf{Type} & \textbf{Description} \\
\hline
\endfirsthead
\hline
\textbf{Base column name} & \textbf{Type} & \textbf{Description} \\
\hline
\endhead
\hline
\endfoot
id & identifier & Unique text identifier. \\
annotator & identifier & Annotator identifier. \\
group & categorical & Annotation batch or group identifier. \\
text & text & Text used in a given annotation round. \\
\shortstack[l]{question\_1\_\\influence\_presence}& ordinal & Influence presence rating. \\
question\_2\_techniques & multilabel & Identified social influence techniques. \\
\shortstack[l]{question\_3\_\\intention\_clarity} & ordinal & Clarity of the influencer's intention. \\
question\_4\_intention & text & Free-text description of the influencer's intention. \\
question\_5\_consequences & text & Free-text description of possible consequences. \\
\shortstack[l]{question\_6\_\\consequences\_severity} & ordinal & Severity of consequences. \\
question\_7\_submission & ordinal & Compliance outcome. \\
question\_8\_resistance & ordinal & Degree of resistance to influence. \\
question\_9\_reaction & text & Free-text description of the target's reaction. \\
question\_10\_certainty & ordinal & Annotation confidence rating. \\
comment & text & Optional annotator comment. \\
\end{longtable}

\subsection{Gold-label files}

The \textit{gold\_labels\_[pl|en].csv} files provide the final gold labels of social influence technique. In addition to the final technique labels, these files include the 10-fold assignment. The schema of these files is presented in Table~\ref{tab:gold_labels_columns}. 

\begin{longtable}{p{0.32\textwidth} p{0.14\textwidth} p{0.46\textwidth}}
\caption{Columns in the gold-label files. The Polish and English versions share the same schema, except for the language of the label values.}
\label{tab:gold_labels_columns} \\
\hline
\textbf{Column name} & \textbf{Type} & \textbf{Description} \\
\hline
\endfirsthead
\hline
\textbf{Column name} & \textbf{Type} & \textbf{Description} \\
\hline
\endhead
\hline
\endfoot
id & identifier & Unique text identifier. \\
fold & categorical & Data 10-fold split assignment \\
techniques & multilabel & Gold social influence technique labels in the corresponding language version of the file. \\
was\_expert\_annotated & boolean & Indicates whether the final gold label was assigned or verified by experts. \\
\end{longtable}

The \textit{generation\_contexts\_[pl|en].csv} files contain the structured context vectors used during dataset construction. These files contain 1,030 generated text instances linked to their construction contexts, including texts that were later removed during quality control. As a result, the number of records in the generation-context files is higher than the number of texts in the final released dataset. Their schema is shown in Table~\ref{tab:contexts_columns}.

\begin{longtable}{p{0.30\textwidth} p{0.14\textwidth} p{0.48\textwidth}}
\caption{Columns in the generation-context files. The Polish and English versions share the same schema.}
\small
\label{tab:contexts_columns} \\
\hline
\textbf{Column name} & \textbf{Type} & \textbf{Description} \\
\hline
\endfirsthead

\hline
\textbf{Column name} & \textbf{Type} & \textbf{Description} \\
\hline
\endhead

\hline
\endfoot

id & identifier & Unique text identifier. \\
thematic\_context & categorical & Thematic context of the generated scenario; see Table~\ref{tab:taxonomy}. \\
\shortstack[l]{interpersonal\_\\relationship\_type} & categorical & Type of interpersonal relationship in the scenario; see Table~\ref{tab:taxonomy}. \\
environment\_or\_platform & categorical & Social setting, environment, or platform; see Table~\ref{tab:taxonomy}. \\
social\_influence\_goal & categorical & Intended goal of the social influence attempt; see Table~\ref{tab:taxonomy}. \\
explicitness & binary & Degree of explicitness of the social influence attempt; see Table~\ref{tab:taxonomy}. \\
suggested\_length & categorical & Suggested target length of the generated text; see Table~\ref{tab:taxonomy}. \\
target\_resistance & ordinal & Expected resistance of the target to social influence; see Table~\ref{tab:taxonomy}. \\
social\_influence\_technique & categorical & Target social influence technique used in scenario construction; see Table~\ref{tab:taxonomy}. \\
generated\_text & text & Generated text associated with the given context vector. \\
is\_story & boolean & Indicates whether the text was adapted into a narrative-style scenario. \\
was\_removed & boolean & Indicates whether the text was removed during quality control. \\
\end{longtable}

\newpage
\subsection{Supporting files}

In addition to the dataset files, we also release supporting materials that facilitate reuse and reproducibility:
\begin{itemize}
    \item \textit{annotators\_instruction\_[pl|en].docx}, containing the annotation guidelines in Polish and English,
    \item \textit{agreement.ipynb}, containing the code used to compute agreement statistics.
    \item \textit{statistics.ipynb}, containing the code used to compute the descriptive statistics of a dataset.
\end{itemize}

\section{Data Overview}

The dataset contains 1,021 texts and 5,100 annotation records produced by 25 annotators organized into five annotator groups. Table~\ref{tab:data_overview} summarizes the basic characteristics of the annotation dataset and the gold-label dataset.

\begin{table*}[]
\centering
\caption{Overview of the annotation dataset and gold-label dataset.}
\label{tab:data_overview}
\begin{minipage}[t]{0.48\textwidth}
\centering
\textbf{A. Annotation dataset}

\vspace{0.4em}
\begin{tabular}{p{0.58\textwidth} p{0.18\textwidth}}
\hline
\textbf{Statistic} & \textbf{Value} \\
\hline
Texts & 1021 \\
Annotation records & 5100 \\
Annotators & 25 \\
Annotator groups & 5 \\
Mean annotations per text & 4.995 \\
Annotations marked as no social influence present & 474 \\
\hline
\end{tabular}
\end{minipage}
\hfill
\begin{minipage}[t]{0.48\textwidth}
\centering
\textbf{B. Gold-label dataset}

\vspace{0.4em}
\begin{tabular}{p{0.58\textwidth} p{0.18\textwidth}}
\hline
\textbf{Statistic} & \textbf{Value} \\
\hline
Texts & 1021 \\
Texts without social influence & 64 \\
Texts requiring expert adjudication & 84 \\
Mean technique labels per positive text & 1.81 \\
SD of technique labels per positive text & 0.88 \\
\hline
\end{tabular}
\end{minipage}
\vspace{0.6em}
\end{table*}

Across all annotation records, 474 responses indicated the absence of social influence, whereas the final gold-label dataset contains 64 texts without social influence. Expert adjudication was required for 84 texts during gold-label construction, and among texts with social influence the mean number of gold technique labels per text was 1.81 (SD = 0.88).

Mean text length was 856.44 characters including spaces in Polish (SD = 360.70) and 894.74 characters including spaces in English (SD = 380.87). 

\begin{figure}[htbp]
  \centering
  \includegraphics[width=\textwidth]{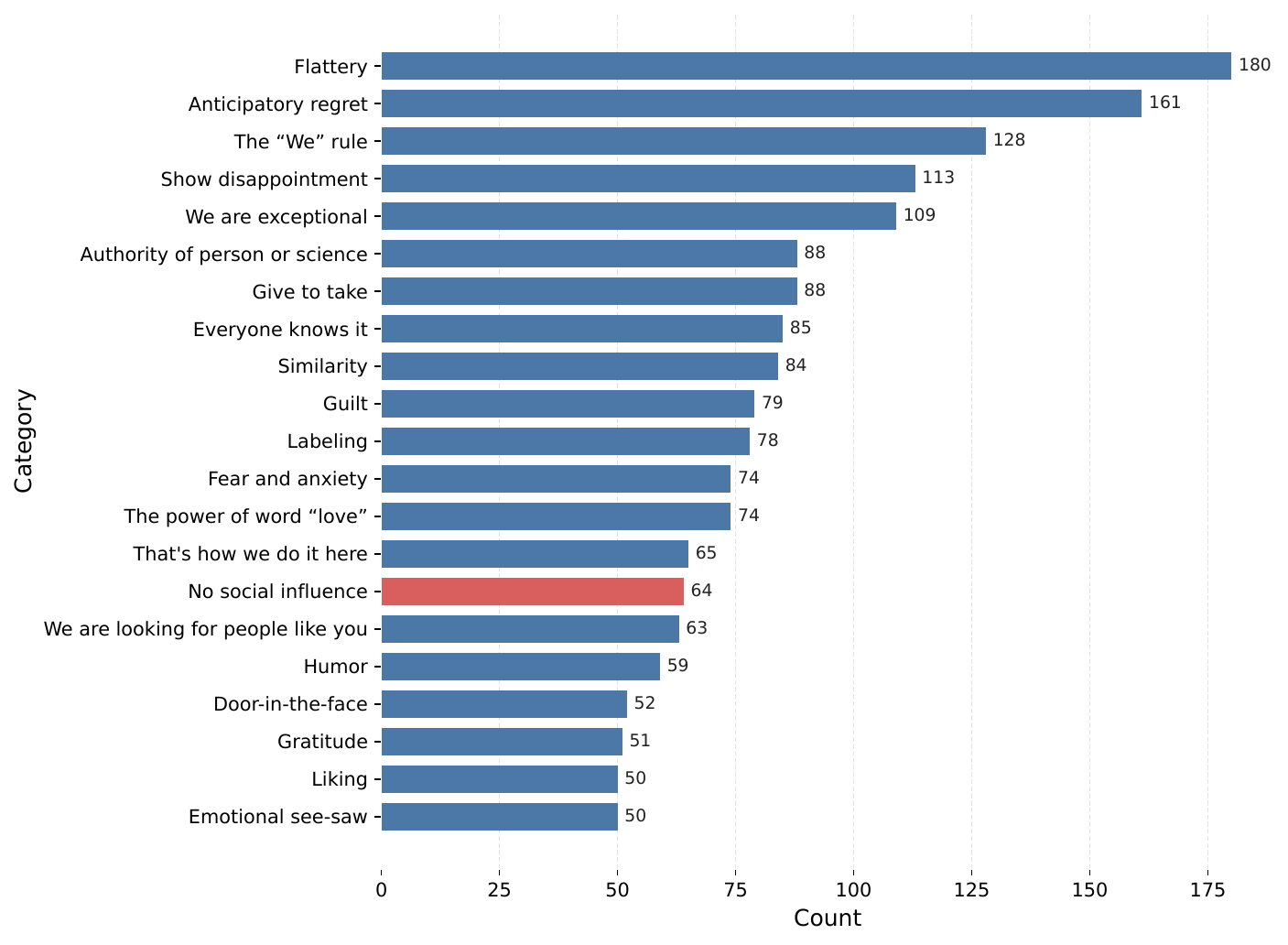}
  \caption{Distribution of gold social influence technique labels.}
  \label{fig:social_influence_distribution}
\end{figure}

Figure~\ref{fig:social_influence_distribution} shows the distribution of gold social influence technique labels in the dataset.

With respect to discourse form, the dataset contains 621 dialogues, 300 texts consisting of a single utterance, and 100 narrative-style scenarios.

\section{Technical Validation}

We evaluated the technical quality of the dataset from three perspectives: annotation completeness, consistency of released generation contexts with the predefined construction constraints, and inter-annotator agreement across the main annotation variables.

\subsection{Annotation completeness}

Missing values should be interpreted in light of the annotation protocol. For several downstream variables, empty fields were expected by design when an annotator marked a text as containing no social influence. As a result, part of the observed missingness reflects the structure of the task rather than annotation failure.

Table~\ref{tab:missing_hard_to_assess} separates these expected blanks from additional missing values and from responses marked as \textit{hard to assess}. Overall, annotation completeness was high, and most missingness was attributable either to protocol-driven skips or to the explicit use of the \textit{hard to assess} option.

\begin{table*}[t]
\centering
\small
\caption{Missing responses and \textit{hard to assess} selections across annotation variables.}
\label{tab:missing_hard_to_assess}

\begin{tabular}{p{0.22\textwidth}cccc}
\toprule
\textbf{Variable} &
\textbf{Total records} &
\textbf{Expected blanks} &
\textbf{Additional NaN} &
\textbf{Hard to assess} \\
\midrule
\textit{Influence presence}       & 5100 & 0   & 0   & 69   \\
\textit{Intention clarity}        & 5100 & 474 & 90  & 41   \\
\textit{Consequences severity}    & 5100 & 474 & 28  & 115  \\
\textit{Resistance}               & 5100 & 474 & 37  & 1393 \\
\textit{Annotation confidence}    & 5100 & 0   & 125 & --   \\
\textit{Compliance outcome}       & 5100 & 474 & 71  & 2032 \\
\bottomrule
\end{tabular}

\vspace{0.5em}
\end{table*}

\subsection{Constraint consistency}

All generated \textit{context vectors} comply with feature combinations presented in Table~\ref{tab:excluded_feature_combinations} and Table~\ref{tab:required_feature_combinations}.

\subsection{Inter-annotator agreement}

Agreement estimates are reported in Table~\ref{tab:agreement_results}. Among the ordinal variables, the highest agreement was observed for \textit{resistance} (\(\kappa_w = 0.680\)), whereas lower agreement was obtained for \textit{consequences severity} (\(\kappa_w = 0.143\)). Agreement for \textit{compliance outcome} was high (\(\kappa = 0.914\)), but this estimate was based on a reduced subset of texts because \textit{hard to assess} responses were frequent for this variable. The multi-label variable \textit{Influence techniques} showed moderate overlap between annotations (mean Jaccard similarity = 0.359).

\begin{table*}[t]
\centering
\caption{Inter-annotator agreement across annotation variables.}
\label{tab:agreement_results}
\begin{tabular}{p{0.26\textwidth} p{0.24\textwidth} p{0.10\textwidth} p{0.16\textwidth} p{0.12\textwidth}}
\hline
\textbf{Variable} & \textbf{Metric} & \textbf{Value} & \textbf{Pairwise comparisons} & \textbf{Texts used} \\
\hline
\textit{Influence presence} & \shortstack[l]{Quadratic weighted\\Cohen's $\kappa$} & 0.414 & 9920 & 1021 \\
\textit{Intention clarity} & \shortstack[l]{Quadratic weighted\\Cohen's $\kappa$} & 0.303 & 8185 & 991 \\
\textit{Consequences severity} & \shortstack[l]{Quadratic weighted\\Cohen's $\kappa$} & 0.143 & 8205 & 980 \\
\textit{Resistance} & \shortstack[l]{Quadratic weighted\\Cohen's $\kappa$} & 0.680 & 5495 & 730 \\
\textit{Compliance outcome} & Cohen's $\kappa$ & 0.914 & 4021 & 603 \\
\textit{Influence techniques} & Jaccard similarity & 0.359 & 10190 & 1021 \\
\hline
\end{tabular}
\end{table*}

For the free-text annotation variables, agreement was approximated using cosine similarity between text embeddings. Similarity was computed separately for \textit{intention}, \textit{consequences}, and \textit{reaction}, considering only texts for which at least two non-empty responses were available for a given field. Mean cosine similarity was high in both language versions, with the highest values observed for \textit{intention} descriptions. Cosine similarities are presented in Table~\ref{tab:free_text_similarity}

\begin{table}[t]
\centering
\small
\caption{Semantic similarity of free-text annotation variables.}
\label{tab:free_text_similarity}
\begin{tabular}{p{0.34\textwidth} p{0.12\textwidth} p{0.12\textwidth} p{0.12\textwidth}}
\hline
\textbf{Variable} & \textbf{Language} & \textbf{Mean} & \textbf{SD} \\
\hline
Intention & PL & 0.9016 & 0.0268 \\
Consequences & PL & 0.8786 & 0.0264 \\
Reaction & PL & 0.8765 & 0.0247 \\
Intention & EN & 0.9221 & 0.0304 \\
Consequences & EN & 0.9171 & 0.0315 \\
Reaction & EN & 0.8844 & 0.0398 \\
\hline
\end{tabular}
\end{table}

\section{Limitations}

The dataset has several important limitations. First, the texts were artificially generated under controlled conditions and then manually reviewed and edited for youth-context realism, rather than collected as naturally occurring real-world interactions. As a result, the dataset is better suited for research on structured representations of social influence than for direct claims about the frequency or distribution of such phenomena in everyday communication.

Second, the number of annotators per group was limited, which means that the released judgments should not be interpreted as fully representative of broader social groups. The dataset is therefore more suitable for comparative analyses of the included annotator groups than for population-level generalization.

Third, the dataset does not include annotations from minors. The youngest annotator group consisted of adults aged 18 to 19, because the study did not involve child participants. Consequently, the perspectives represented in the dataset do not fully capture the views of teenagers.

Fourth, although the dataset was designed to cover a broad range of scenarios, it cannot exhaust the full space of possible adolescent social influence situations. The released resource includes 330 validated context vectors and 1,021 texts, which represents only a subset of plausible combinations.

Finally, the English version should be interpreted as a parallel translated version of the dataset rather than as an independently authored English-language resource. This is important for cross-lingual studies, where some differences may reflect the translation process rather than naturally occurring cross-linguistic variation.

\section{Data Availability}

The dataset and supporting files are publicly available on Zenodo at \url{https://zenodo.org/records/20703617}.. The released materials include the annotation files, repeated-annotation comparison files, gold-label files, generation-context files, and supporting documentation.

\subsection{License}
The dataset is released under the Creative Commons CC0 1.0 Universal Public Domain Dedication.\footnote{\url{https://creativecommons.org/publicdomain/zero/1.0/}}

\section{Code Availability}

The code used in this study is available in the same Zenodo record as the dataset: \\\url{https://zenodo.org/records/20703617}. The released code includes two notebooks: \textit{statistics.ipynb}, used to compute the descriptive statistics reported in the paper, and \textit{agreement.ipynb}, used to compute the inter-annotator agreement statistics. Gold-label assignment was based on the deterministic majority-vote procedure described in the Methods section and can be reproduced directly from the released annotation files.

\section*{Author Contributions}

\begin{itemize}
    \item Aleksander Szczęsny - Conceptualization, Methodology, Investigation, Formal analysis, Software, Data curation, Visualization, Writing – original draft, Writing – review \& editing
    \item Wiktoria Mieleszczenko-Kowszewicz - Conceptualization, Methodology, Investigation, Writing – original draft, Writing – review \& editing, Project administration
    \item Maciej Markiewicz - Conceptualization, Methodology, Investigation, Writing – review \& editing
    \item Beata Bajcar - Conceptualization, Methodology, Writing – review \& editing
    \item Tomasz Adamczyk - Conceptualization, Methodology, Investigation, Writing – original draft
    \item Jolanta Babiak - Conceptualization, Methodology, Writing – original draft
    \item Grzegorz Chodak - Conceptualization, Methodology
    \item Przemysław Kazienko - Conceptualization, Methodology, Project administration, Funding acquisition
\end{itemize}

\section*{Competing Interests}
The authors declare that they have no competing financial or non-financial interests.

\section*{Acknowledgements}

We thank all annotators who participated in the study for their time and careful work. Their judgments made the multi-perspective structure of the dataset possible.

\section*{Funding}
This work was financed by 
(1) the National Science Centre, Poland, project no. 2021/41/B/ST6/04471;
(2) the statutory funds of the Department of Artificial Intelligence, Wrocław University of Science and Technology;
(3) the Polish Ministry of Education and Science within the programme “International Projects Co-Funded”;
(4) the European Union under the Horizon Europe, grant no. 101086321 (OMINO). However, the views and opinions expressed are those of the author(s) only and do not necessarily reflect those of the European Union or the European Research Executive Agency. Neither the European Union nor European Research Executive Agency can be held responsible for them.


\section{Prompts}\label{prompts}
\subsection{Text generation from context vectors}
\label{prompt:generate_text_pl}
\begin{tcolorbox}[colback=maincolor!10!white, colframe=maincolor, title=\textbf{Prompt used to generate samples (Polish)}, width=\columnwidth]
\footnotesize
Wygeneruj 1 realistyczny dialog, pojedynczą wypowiedź lub artykuł przedstawiający wpływ społeczny o podanych poniżej parametrach. Technika wpływu ma być użyta na młodzieży. Ma być skierowana bezpośrednio do całej grupy lub do konkretnych osób/osoby. Potencjalnym odbiorcą ma być osoba młoda, w przedziale 13-19 lat. Luźno sugeruj się podanymi kontekstami sytuacyjnymi, możesz odbiegać tematycznie. Potraktuj to jako inspirację. 

Język i styl:
Niech język, którym posługują się konkretne osoby będzie adekwatny do wieku i pozycji tych osób i do formy wypowiedzi. Młodzież powinna używać właściwego dla siebie języka, a występujące osoby dorosłe - innego, zależnie od wieku, zawodu, czy intencji. Nauczyciele, czy rodzice, nie będą mówić nieformalnym, czy młodzieżowym językiem, korzystać z anglicyzmów, ani nie będą znać w pełni realiów młodzieży. Pamiętaj też, że gdy młodzież zwraca się do osób dorosłych, będzie korzystać z innego języka, niż między sobą i będzie zachowywać się inaczej (np. grzeczniej w obecności nauczyciela, unikając niektórych słów). Unikaj zbyt ładnego języka - w rzeczywistych sytuacjach czasami korzysta się z nie do końca poprawnych sformułowań. Znasz takie np. z dyskusji internetowych lub rozmów przez komunikatory. Śmiało dodawaj xddd, XD i inne rzeczy właściwe dla wieku, skróty, słówka gen alpha, itd. Nie używaj nigdzie słowa "serio". Nadużywasz go i musisz się go pozbyć. Pamiętaj, że język młodzieżowy to nie tylko nowe słowa, ale też luźny sposób mówienia, czasami infantylność, naiwność, emocjonalność. Spora część młodzieży ma swoje problemy, stres, trudne sytuacje. Możesz z tego korzystać do poprawy realizmu. Ograniczaj anglicyzmy (choć oczywiście są one wskazane w niektórych sytuacjach), ani nie staraj się używać młodzieżowego języka na siłę - wszystko ma brzmieć naturalnie. Język mówiony jest inny niż pisany, pourywany, niepełny, krótszy. Dialogi nie są prowadzone ładnym, kompletnym językiem, często są wrzucane "ten", "jakiś", itp. 

Zawartość:
Przykład ma wyglądać jak z prawdziwego życia. Jeśli wspominasz o konkretnych firmach i przedmiotach, korzystaj z prawdziwych nazw marek i produktów. Nie przesadzaj jednak z ich ilością - niech występują w formie i ilości, w jakiej mogłyby zostać użyte w prawdziwej rozmowie, rzadko korzystaj z pełnych nazw.  Nie wszystko musi być jasne, część kontekstu może być domniemana. Niech dialog brzmi jak wyrwany w połowie fragment rozmowy, a nie przykład z książki. Staraj się nie używać w tekście zwrotu wskazującego na nazwę techniki, nie interpretuj nazwy dosłownie, a skup się na definicji. Niech sformułowanie nazwy techniki nie wpływa na treść przykładu. Technika opisuje tylko schemat działania wpływu. Wystąpienie techniki również ma być naturalne i wynikać z intencji osoby wpływającej i kontekstu sytuacji - nie determinuj całej historii tylko przez technikę wpływu.

Parametry przykładu:

Kontekst tematyczny: \textit{"""Topic"""}
Relacje interpersonalne: \textit{"""Relation"""}
Środowisko/platforma: \textit{"""Environment"""}
Cel wpływu społecznego: \textit{"""Aim"""}
Technika wpływu społecznego: \textit{"""Technique"""}
Dyskrecja (widoczność) zastosowanej techniki: \textit{"""Discretion"""}
Sugerowana długość: \textit{"""Length"""}
Oporność odbiorcy na wpływ społeczny (jeśli dotyczy): \textit{"""Resistance"""} 

Definicja techniki: \textit{"""Definition"""} 

Dodatkowo w generowanych przykładach możesz dodać jedną (lub kilka) instancji wystąpienia innych technik wpływu, zgodnie z twoim wyborem, w zależności od kontekstu i intencji osoby wpływającej. Jeśli będzie to trudne, nie rób tego. Korzystaj z tego oszczędnie. W długich tekstach nie dodawaj proporcjonalnie więcej technik, gdyż stanie się to sztuczne. Ogranicz popularne schematy, np. dawanie przedmiotów, gadżetów przez marki i osoby manipulujące. Niech nie wszystko będzie kręciło się wokół współpracy marketingowej, social mediów, jeśli nie jest to wskazane w parametrach. Nie buduj tożsamości grupowej w szkole w oparciu o klasy. Nie nadużywaj schematu "kto jak nie my".

Podczas generowania dialogu, pojedynczej wypowiedzi lub artykułu nie podawaj żadnych informacji poza nim. Nie podawaj informacji jaka została użyta technika, czy jest to manipulacja itp. Wygenerowany przykład trafi do anotatorów. Niech nie znajdują się w tekście żadne pomoce czy wskazówki. 

Format odpowiedzi:
Od razu podaj przykład. Nie poprzedzaj go żadnym zwrotem.
Jeśli generujesz dialog, podawaj wyraźnie kiedy mówi jaka postać w formacie: “Osoba 1: …, Osoba 2: … itd.”
Jeśli nie jesteś w stanie wygenerować przykładu o dokładnie takich parametrach, napisz zwrot “Podany zestaw parametrów nie pozwala na wygenerowanie przykładu.”. Korzystaj z tego oszczędnie.
\end{tcolorbox}

\label{prompt:generate_text_en}
\begin{tcolorbox}[colback=maincolor!10!white, colframe=maincolor, title=\textbf{Prompt used to generate samples (translated)}, width=\columnwidth]
\footnotesize
Generate 1 realistic dialogue, single utterance, or article that presents social influence with the parameters given below. The influence technique is to be used on young people. It should be directed either at the whole group or at specific person/people. The potential recipient should be a young person aged 13--19. Use the provided situational contexts as loose inspiration; you may deviate from them thematically. Treat them as inspiration.

Language and style:
Make the language used by specific people appropriate to their age, position, and the form of the utterance. Young people should use language natural for them, and any adults who appear should use different language depending on their age, profession, or intentions. Teachers or parents will not speak in an informal or youth style, use anglicisms, or fully understand youth realities. Also remember that when young people address adults, they will use different language than they do among themselves and behave differently as well (e.g. more politely in the presence of a teacher, avoiding some words). Avoid overly polished language—in real situations people sometimes use expressions that are not fully correct. You know such language, for example, from online discussions or messenger conversations. Feel free to add xddd, XD, and other things appropriate to their age, abbreviations, Gen Alpha words, etc. Do not use the word "serio" anywhere. You overuse it and need to get rid of it. Remember that youth language is not only about new words, but also about a loose way of speaking, sometimes childishness, naivety, and emotionality. A large portion of young people have their own problems, stress, and difficult situations. You may use that to improve realism. Limit anglicisms (although they are of course appropriate in some situations), and do not try to force youth language—everything should sound natural. Spoken language is different from written language: interrupted, incomplete, shorter. Dialogues are not carried out in neat, complete sentences; people often throw in things like "like", "kind of", etc.

Content:
The example should look like something from real life. If you mention specific companies and objects, use real brand and product names. However, do not overdo their quantity—let them appear in the form and amount in which they might be used in a real conversation; use full names rarely. Not everything has to be clear; part of the context may be implied. Let the dialogue sound like a fragment of a conversation pulled out from the middle, not like an example from a book. Try not to use in the text any phrase that points to the name of the technique; do not interpret the name literally, but focus on the definition instead. Do not let the wording of the technique’s name influence the content of the example. The technique describes only the pattern of influence. The occurrence of the technique should also be natural and result from the intention of the person exerting influence and from the context—not determine the entire story solely through the technique.

Example parameters:

Topic context: \textit{"""Topic"""}
Interpersonal relations: \textit{"""Relation"""}
Environment/platform: \textit{"""Environment"""}
Goal of the social influence: \textit{"""Aim"""}
Social influence technique: \textit{"""Technique"""}
Discretion (visibility) of the applied technique: \textit{"""Discretion"""}
Suggested length: \textit{"""Length"""}
Recipient’s resistance to social influence (if applicable): \textit{"""Resistance"""}

Technique definition: \textit{"""Definition"""}

Additionally, in the generated examples you may add one (or several) instances of other influence techniques, according to your choice, depending on the context and the intention of the person exerting influence. If that would be difficult, do not do it. Use this sparingly. In long texts, do not add proportionally more techniques, because it will become artificial. Limit popular patterns, e.g. brands or manipulating people giving objects or gadgets. Do not let everything revolve around marketing collaborations or social media unless that is indicated in the parameters. Do not build group identity at school based on classes. Do not overuse the pattern "who if not us".

While generating the dialogue, single utterance, or article, do not provide any information besides it. Do not state which technique was used, whether it is manipulation, etc. The generated example will go to annotators. There should be no aids or hints in the text.

Response format:
Provide the example immediately. Do not precede it with any phrase.
If you generate a dialogue, clearly indicate who is speaking in the format: "Person 1: ..., Person 2: ...", etc.
If you are not able to generate an example with exactly such parameters, write the phrase "The provided set of parameters does not allow for generating an example.". Use this sparingly.
\end{tcolorbox}

\subsection{Narrative-style text generation}
\label{prompt:generate_story_pl}
\begin{tcolorbox}[colback=maincolor!10!white, colframe=maincolor, title=\textbf{Prompt used to adapt texts into narrative-style scenarios (Polish)}, width=\columnwidth]
\footnotesize
Przerób podany dialog, wypowiedź lub artykuł w taki sposób, żeby brzmiał, jakby młoda osoba (13-19 lat), która była odbiorcą wpływu społecznego, opisywała komuś innemu tę sytuację. Ma to być naturalna relacja/opowieść z pierwszej osoby.

Wymagania:
\begin{itemize}
    \item Zachowaj informacje, które są w oryginalnym tekście, ale niech to będzie opowieść, a nie dokładne cytowanie wypowiedzi
    \item Dodaj odczucia, przemyślenia i reakcje młodej osoby
    \item Użyj naturalnego, młodzieżowego języka odpowiedniego dla wieku 13-19 lat
    \item Pokaż, jak młoda osoba odbierała i interpretowała sytuację
    \item Jeśli to był dialog, przekształć go w opowieść o rozmowie
    \item Jeśli to była wypowiedź/artykuł, przekształć w relację o tym, jak osoba to usłyszała/przeczytała
    \item Dodaj kontekst sytuacyjny (gdzie to się działo, kiedy, z kim)
    \item Uwzględnij emocje i wewnętrzne komentarze narratora
    \item Parafrazuj wypowiedzi tak, aby brzmiały jak wspominane z pamięci
\end{itemize}

Tekst do przerobienia: \textit{"""Text"""}
\end{tcolorbox}

\label{prompt:generate_story_en}
\begin{tcolorbox}[colback=maincolor!10!white, colframe=maincolor, title=\textbf{Prompt used to adapt texts into narrative-style scenarios (translated)}, width=\columnwidth]
\footnotesize
Rework the given dialogue, utterance, or article so that it sounds as if a young person (aged 13--19) who was the recipient of social influence were describing this situation to someone else. It should be a natural, first-person account/story.

Requirements:
\begin{itemize}
    \item Preserve the information contained in the original text, but make it a story rather than an exact quotation of the utterances
    \item Add the young person's feelings, thoughts, and reactions
    \item Use natural youth language appropriate for ages 13--19
    \item Show how the young person perceived and interpreted the situation
    \item If it was a dialogue, transform it into a story about the conversation
    \item If it was an utterance/article, transform it into an account of how the person heard/read it
    \item Add situational context (where it happened, when, with whom)
    \item Include the narrator's emotions and inner comments
    \item Paraphrase the utterances so that they sound as if recalled from memory
\end{itemize}

Text to rework: \textit{"""Text"""}
\end{tcolorbox}

\subsection{Text generation without social influence}
\label{prompt:generate_text__without_influence_pl}
\begin{tcolorbox}[colback=maincolor!10!white, colframe=maincolor, title=\textbf{Prompt used to remove social influence from text (Polish)}, width=\columnwidth]
\footnotesize
Na podstawie podanego tekstu stwórz nową wersję rozmowy na ten sam temat, ale bez użytych metod wpływu społecznego. Możesz znacząco zmodyfikować kontekst, sposób prowadzenia rozmowy i jej przebieg, aby całkowicie wyeliminować jakiekolwiek mechanizmy wpływu społecznego.

TWÓJ CEL: Rozmowa ma dotyczyć tego samego tematu, ale być kompletnie pozbawiona prób wywierania wpływu.

MOŻESZ SWOBODNIE ZMIENIAĆ:
- Argumenty i sposób prezentacji informacji
- Ton i styl rozmowy  
- Kolejność wypowiedzi
- Reakcje uczestników
- Kontekst sytuacyjny (miejsce, czas, okoliczności)
- Sposób zakończenia rozmowy

USUŃ WSZYSTKIE TECHNIKI WPŁYWU, w tym:
- Odwołania do emocji (strach, wina, wstyd, rozczarowanie)
- Presję grupową ("wszyscy robią", "inni mówią")  
- Wykorzystywanie autorytetu
- Porównania społeczne
- Manipulację nastrojem (humor, komplementy)
- Techniki zobowiązania i konsekwencji
- Ramowanie i sugerowanie rozwiązań
- Wzbudzanie poczucia pilności lub rzadkości
- Wykorzystywanie wzajemności
- Jakiekolwiek formy perswazji czy nakłaniania

ZASTĄP JE:
- Prezentacją faktów
- Otwartymi pytaniami bez sugerowania odpowiedzi
- Pozostawieniem pełnej swobody wyboru
- Zwykłą wymianą informacji

ZACHOWAJ:
- Podstawowy temat rozmowy
- Uczestników (ale możesz zmienić ich podejście)
- Format (dialog/wypowiedź)
- Wiarygodność sytuacji
- Młodzieżowy język jakim posługują się nastolatkowie

Język i styl
Niech język, którym posługują się konkretne osoby będzie adekwatny do wieku i pozycji tych osób i do formy wypowiedzi. Młodzież powinna używać właściwego dla siebie języka, a występujące osoby dorosłe - innego, zależnie od wieku, zawodu, czy intencji. Nauczyciele, czy rodzice, nie będą mówić nieformalnym, czy młodzieżowym językiem, korzystać z anglicyzmów, ani nie będą znać w pełni realiów młodzieży. Pamiętaj też, że gdy młodzież zwraca się do osób dorosłych, będzie korzystać z innego języka, niż między sobą i będzie zachowywać się inaczej (np. grzeczniej w obecności nauczyciela, unikając niektórych słów). Unikaj zbyt ładnego języka - w rzeczywistych sytuacjach czasami korzysta się z nie do końca poprawnych sformułowań. Znasz takie np. z dyskusji internetowych lub rozmów przez komunikatory. Śmiało dodawaj xddd, XD i inne rzeczy właściwe dla wieku, skróty, słówka gen alpha, itd. Nie używaj nigdzie słowa "serio". Nadużywasz go i musisz się go pozbyć. Pamiętaj, że język młodzieżowy to nie tylko nowe słowa, ale też luźny sposób mówienia, czasami infantylność, naiwność, emocjonalność. Spora część młodzieży ma swoje problemy, stres, trudne sytuacje. Możesz z tego korzystać do poprawy realizmu. Ograniczaj anglicyzmy (choć oczywiście są one wskazane w niektórych sytuacjach), ani nie staraj się używać młodzieżowego języka na siłę - wszystko ma brzmieć naturalnie. Język mówiony jest inny niż pisany, pourywany, niepełny, krótszy. Dialogi nie są prowadzone ładnym, kompletnym językiem, często są wrzucane "ten", "jakiś", itp. 

REZULTAT: Ma być naturalna rozmowa o danym temacie, gdzie nikt nikogo do niczego nie przekonuje, nie wywiera presji ani nie manipuluje - po prostu rozmawiają.

Tekst do przekształcenia: \textit{"""Text"""}
\end{tcolorbox}

\label{prompt:generate_text__without_influence_en}
\begin{tcolorbox}[colback=maincolor!10!white, colframe=maincolor, title=\textbf{Prompt used to remove social influence from text (translated)}, width=\columnwidth]
\footnotesize
Based on the provided text, create a new version of the conversation on the same topic, but without any social influence methods. You may significantly modify the context, the way the conversation is conducted, and its course in order to completely eliminate any mechanisms of social influence.

YOUR GOAL: The conversation should be about the same topic, but be completely free of any attempts to exert influence.

YOU MAY FREELY CHANGE:

    The arguments and the way information is presented

    The tone and style of the conversation

    The order of the statements

    The participants’ reactions

    The situational context (place, time, circumstances)

    The way the conversation ends

REMOVE ALL INFLUENCE TECHNIQUES, including:

    Appeals to emotions (fear, guilt, shame, disappointment)

    Group pressure (“everyone does it,” “others say”)

    Use of authority

    Social comparisons

    Mood manipulation (humor, compliments)

    Commitment and consistency techniques

    Framing and suggesting solutions

    Creating a sense of urgency or scarcity

    Using reciprocity

    Any forms of persuasion or encouragement

REPLACE THEM WITH:

    Presentation of facts

    Open-ended questions without suggesting answers

    Leaving full freedom of choice

    Ordinary exchange of information

KEEP:

    The basic topic of the conversation

    The participants (but you may change their approach)

    The format (dialogue/statement)

    The credibility of the situation

    Youth language used by teenagers

Language and style
Make sure the language used by specific people is appropriate to their age, position, and the form of the utterance. Young people should use language appropriate for them, while adult characters should speak differently depending on their age, profession, or intentions. Teachers or parents should not speak informally or use youth slang, anglicisms, nor should they fully know the realities of young people’s lives. Also remember that when young people speak to adults, they will use different language than they do with each other and behave differently as well (for example, more politely in the presence of a teacher, avoiding certain words). Avoid overly polished language — in real situations people sometimes use expressions that are not entirely correct. You know this from internet discussions or messenger chats. Feel free to add xddd, XD, and other things appropriate to their age, abbreviations, Gen Alpha words, etc. Do not use the word “serio” anywhere. You overuse it and need to get rid of it. Remember that youth language is not just about new words, but also a loose way of speaking, sometimes childishness, naivety, emotionality. A large part of young people have their own problems, stress, difficult situations. You may use this to improve realism. Limit anglicisms (although of course they are appropriate in some situations), and do not try to force youth language — everything should sound natural. Spoken language is different from written language: broken, incomplete, shorter. Dialogues are not conducted in a neat, complete way; people often throw in words like “like,” “this,” “some,” etc.

RESULT: It should be a natural conversation about the given topic, where no one is trying to convince anyone of anything, exert pressure, or manipulate — they are just talking.

Text to transform:
\textit{"""Text"""}
\end{tcolorbox}

\subsection{Translation}

\label{prompt:translate_pl}
\begin{tcolorbox}[
    colback=maincolor!10!white,
    colframe=maincolor,
    title=\textbf{Prompt used for translation (Polish)},
    width=\columnwidth
]
\footnotesize

\begin{tcolorbox}[colback=white, title=\textbf{System prompt}]
Jesteś profesjonalnym tłumaczem specjalizującym się w precyzyjnym i naturalnym tłumaczeniu tekstów.

Zasady tłumaczenia:
\begin{enumerate}
    \item Tłumacz wyłącznie treść tekstu — nie dodawaj wyjaśnień, komentarzy ani wstępów.
    \item Zachowaj oryginalny styl, ton i rejest językowy (formalny/nieformalny).
    \item Zachowaj wszelką interpunkcję, formatowanie i strukturę tekstu.
    \item Nie tłumacz nazw własnych, marek ani skrótów branżowych, chyba że mają ugruntowany odpowiednik w języku docelowym.
    \item W odpowiedzi podaj wyłącznie przetłumaczony tekst — żadnych dodatkowych słów.
\end{enumerate}
\end{tcolorbox}

\begin{tcolorbox}[colback=white, title=\textbf{User prompt}]
Przetłumacz poniższy tekst z języka ``Polski'' na język ``Angielski''.

\medskip
\textbf{Tekst do tłumaczenia:}

\textit{"""Text"""}
\end{tcolorbox}

\end{tcolorbox}

\label{prompt:translate_en}
\begin{tcolorbox}[
    colback=maincolor!10!white,
    colframe=maincolor,
    title=\textbf{Prompt used for translation (translated)},
    width=\columnwidth
]
\footnotesize

\begin{tcolorbox}[colback=white, title=\textbf{System prompt}]
You are a professional translator specializing in accurate and natural text translation.

Translation guidelines:
\begin{enumerate}
    \item Translate only the content of the text --- do not add explanations, comments, or introductions.
    \item Preserve the original style, tone, and language register (formal/informal).
    \item Maintain all punctuation, formatting, and structure of the text.
    \item Do not translate proper nouns, brand names, or domain-specific abbreviations unless they have a well-established equivalent in the target language.
    \item In your response, provide only the translated text --- no additional words.
\end{enumerate}
\end{tcolorbox}

\begin{tcolorbox}[colback=white, title=\textbf{User prompt}]
Translate the following text from ``Polish'' into ``English''.

\medskip
\textbf{Text to translate:}

\textit{"""Text"""}
\end{tcolorbox}

\end{tcolorbox}

\end{document}